\documentclass[10pt,twocolumn,letterpaper]{article}

\usepackage{iccv}
\usepackage{times}
\usepackage{epsfig}
\usepackage{graphicx}
\usepackage{amsmath}
\usepackage{amssymb}
\usepackage{makecell}
\usepackage{bm}

\usepackage{xcolor}
\usepackage{soul}

\usepackage{fancyvrb}
\usepackage{algorithm}
\usepackage{algorithmic}
\usepackage{subfig}

\graphicspath{{figs/}}
\newcommand{\bx}{\mathbf{x}}
\newcommand{\bz}{\mathbf{z}}
\newcommand{\bT}{\mathbf{T}}

\usepackage[pagebackref=true,breaklinks=true,letterpaper=true,colorlinks,bookmarks=false]{hyperref}

\iccvfinalcopy 

\def\iccvPaperID{14} 

\ificcvfinal\fi

\begin{document}

\title{Satellite Pose Estimation with Deep Landmark Regression\\and Nonlinear Pose Refinement}

\author{Bo Chen, Jiewei Cao, \'{A}lvaro Parra, Tat-Jun Chin\\
School of Computer Science, The University of Adelaide\\
Adelaide, South Australia, 5005 Australia\\
{\tt\small \{bo.chen, jiewei.cao, alvaro.parrabustos, tat-jun.chin\}@adelaide.edu.au}
}

\maketitle
\ificcvfinal\thispagestyle{empty}\fi

\begin{abstract}
We propose an approach to estimate the 6DOF pose of a satellite, relative to a canonical pose, from a single image. Such a problem is crucial in many space proximity operations, such as docking, debris removal, and inter-spacecraft communications. Our approach combines machine learning and geometric optimisation, by predicting the coordinates of a set of landmarks in the input image, associating the landmarks to their corresponding 3D points on an a priori reconstructed 3D model, then solving for the object pose using non-linear optimisation. Our approach is not only novel for this specific pose estimation task, which helps to further open up a relatively new domain for machine learning and computer vision, but it also demonstrates superior accuracy and won the first place in the recent Kelvins Pose Estimation Challenge organised by the European Space Agency (ESA).
\end{abstract}

\section{Introduction}\label{sec:intro}


Estimating the 6DOF pose of space-borne objects (e.g., satellites, spacecraft, orbital debris) is a crucial step in many space operations such as docking, non-cooperative proximity tasks (e.g., debris removal), and inter-spacecraft communications (e.g., establishing quantum links). Existing solutions are mainly based on active sensor-based systems, e.g., the TriDAR system which uses LiDAR~\cite{English2005tridar, Ruel2012space}. Recently, monocular pose estimation techniques for space applications are drawing significant attention due to their lower power consumption and relatively simple requirements~\cite{D2014pose, Sharma2018robust, Sharma2019pose, reviewPAS19}.

\begin{figure}[h]
    \centering
    \includegraphics[width = \linewidth]{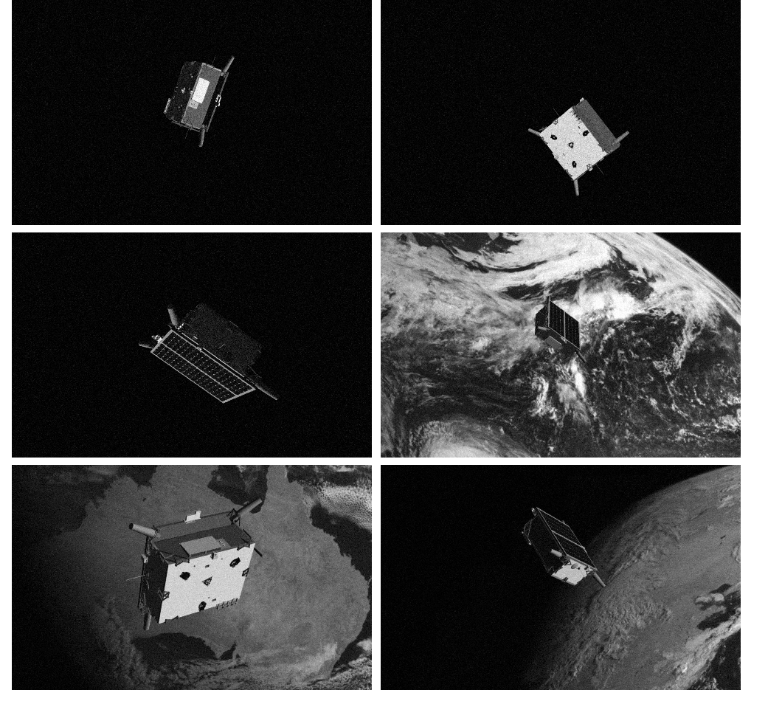}
    \caption{Sample images of the Tango satellite from SPEED~\cite{Sharma2019pose}. Note the significant variations in object size, object orientation, background and lighting condition.}
    \label{fig:montage}
\end{figure}



\begin{figure*}[h]
    \centering
    \includegraphics[width = \linewidth]{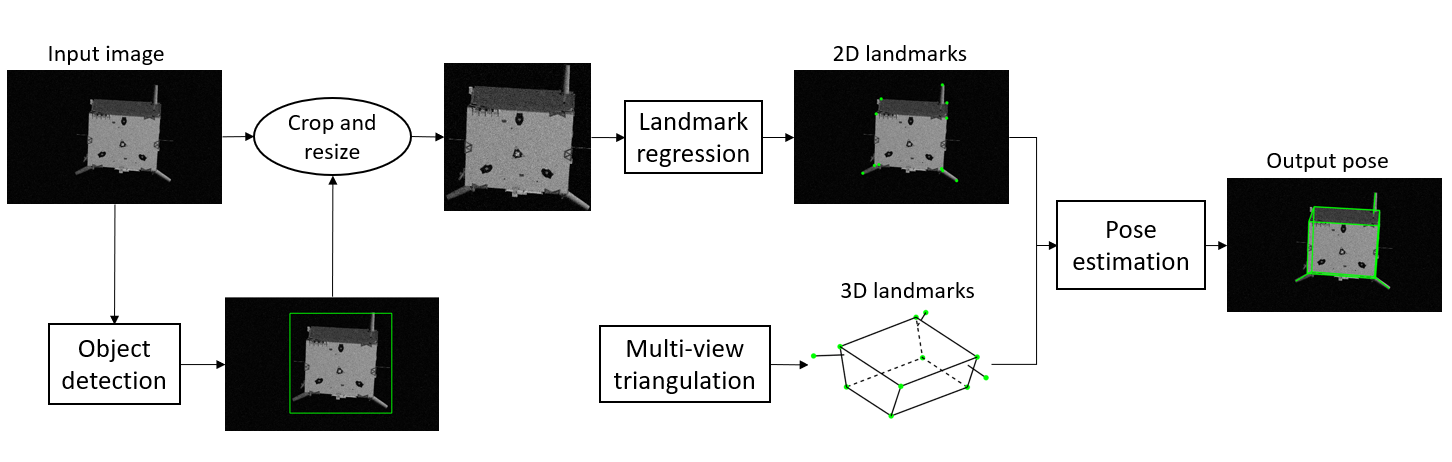}
    \caption{Overall pipeline of our satellite pose estimator.}
    \label{fig:pipeline}
\end{figure*}

Due to the importance of the problem, the Advanced Concepts Team (ACT) at ESA recently held a benchmark competition called Kelvins Pose Estimation Challenge (KPEC)~\cite{kelvins}; given images that depict a known satellite under different unknown poses (see Figure~\ref{fig:montage}), estimate the pose of the satellite in each image. To develop their algorithms, the challenge participants are given a set of training images containing the target satellite with ground truth poses; Section~\ref{sec:dataset} provides more details of the dataset.

The scenario considered in KPEC is a special case of monocular vision-based object pose estimation~\cite{hodan2018bop, Sundermeyer_2018_ECCV}. This is because the target object (the ``Tango" satellite) is known beforehand, and there is no need to generalize the pose estimator to unseen-before instances of the object class (e.g., other satellites). However, the background environment can still vary, as exemplified in Figure~\ref{fig:montage}. Contrast the KPEC scenario to the generic pose estimation setting~\cite{hodan2018bop, Sundermeyer_2018_ECCV}, where the provenance of the target object is unknown \emph{a priori} and generalising to unseen-before instances is necessary (e.g., a car pose estimator must work on all kinds of cars).


Under the KPEC setting, we developed a monocular pose estimation technique for space-borne objects such as satellites. Inspired by works that combine the strength of deep neural networks and geometric optimisation~\cite{Peng2019pvnet, Pavlakos20176,Tekin2018real}, our approach contains three main components:
\begin{enumerate}
    \item using the training images, reconstruct a 3D model of the satellite by multi-view triangulation;
    \item train a deep network to predict the position of pre-defined landmark points in the input image;
    \item solve for the pose of the object in the image using the 2D-3D correspondences of the predicted landmarks via robust geometric optimisation.
\end{enumerate}
A high level pipeline of our framework is illustrated in Figure~\ref{fig:pipeline}. Our code can be accessed in~\cite{spe}.

As suggested above, our method fully takes advantages of all available data and assumptions of the problem. This plays a significant role in producing highly-accurate 6DOF pose estimation for the KPEC. Specifically, our method commits an average cross validation (CV) error of 0.7277 degrees for orientation and 0.0359 metres for translation on the KPEC training set. We achieved an overall score of 0.0094 on the test set which ranked us the first place in KPEC. The rest of the paper first reivews related works and then describes our method and results in detail.

\subsection{Dataset}\label{sec:dataset}

\begin{figure}[b]
    \centering
    \includegraphics[width = \linewidth]{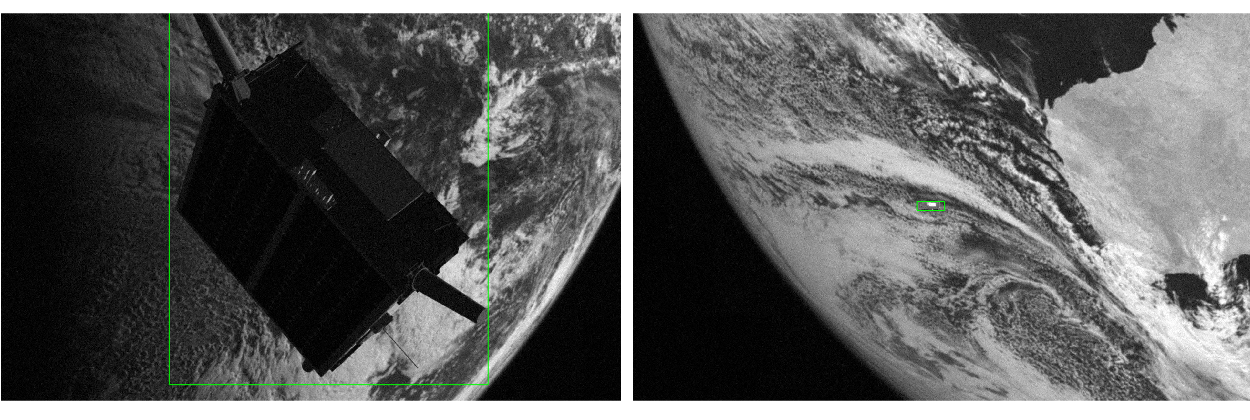}
    \caption{Large variation in object size in the images.}
    \label{fig:big_small}
\end{figure}

The KPEC was designed around the Spacecraft PosE Estimation Dataset (SPEED) \cite{Sharma2019pose}, which consists of high-fidelity grayscale images of the Tango satellite; see Figure~\ref{fig:montage}. There are 12,000 training images with ground truth 6DOF poses (position and orientation) and 2,998 testing images without ground truth. Each image is of size 1920$\times$1200 pixels. Half of the available images have no background (i.e., the background is the space void) while the other half contain the Earth as the background. Mirroring the setting during proximity operations, the size, orientation and lighting condition of the satellite in the images vary significantly, e.g., the number of object pixels vary between 1k and 500k; see Figure~\ref{fig:big_small} for an example. For more details of the dataset, see~\cite{Sharma2019pose}. 

\begin{figure*}[ht!]
    \centering
    \includegraphics[width = 0.95\linewidth]{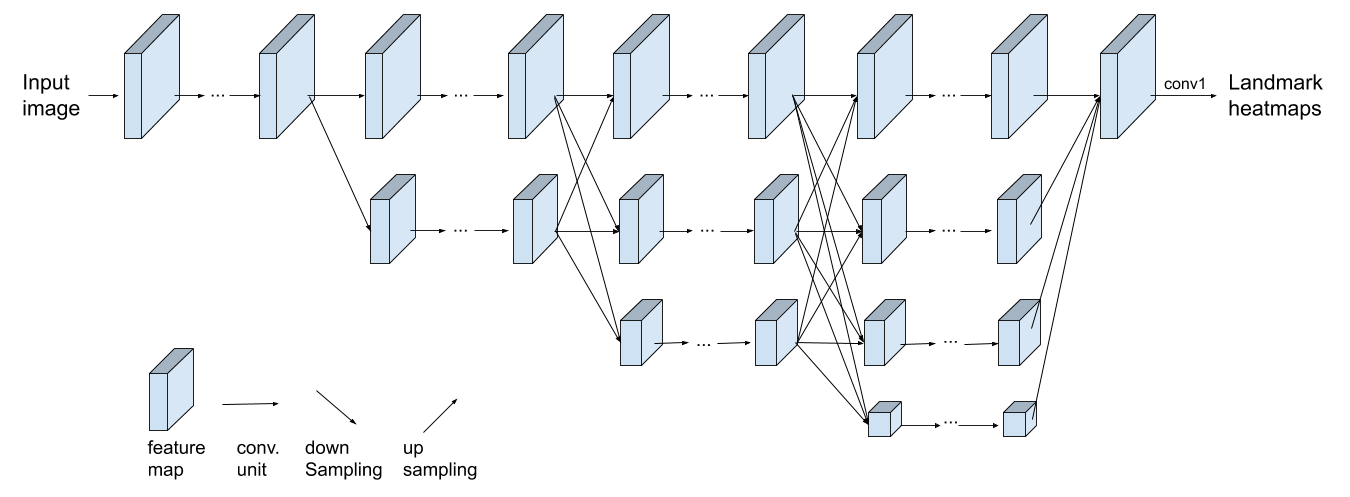}
    \caption{Illustration of the HRNet architecture in our landmark  regression model.}
    \label{fig:hrnet}
\end{figure*}

\section{Related works}\label{sec:related}

Monocular vision-based pose estimation has a large body of literature. We review the major classes of previous work, before surveying the specific case of spacecraft pose estimation.

\subsection{Monocular pose estimation}

\paragraph{Keypoint methods}

Traditional pose estimation techniques usually use hand-crafted keypoint detectors and descriptors, e.g., SIFT~\cite{Lowe1999object, Lowe2004distinctive}, SURF~\cite{Bay2006surf}, MSER~\cite{Matas2004robust} and BRIEF~\cite{Calonder2010brief}. The key step is to produce a set of 2D-2D or 2D-3D keypoint correspondences, then estimate the pose using non-linear optimisation from the correspondence set. The keypoints are detected automatically and described using heuristic measures of geometric and photometric invariance. However, while the keypoint methods are robust to a certain extent, they typically fail where there is large variations in pose and lighting conditions. Nonetheless, the earlier research has given birth to effective and well-understood geometric algorithms (e.g., PnP solvers) that are able to estimate the pose accurately and robustly, \emph{given} a reasonable correspondence set; we exploit these techniques in our pipeline.

\paragraph{End-to-end learning}

The success of deep learning in image classification and object detection has motivated a large number of works on end-to-end learning for pose estimation~\cite{Sundermeyer_2018_ECCV,Brahmbhatt2018geometry,Kendall2016modelling,Kendall2017geometric,Kendall2015posenet,Melekhov2017image}. Generally speaking, these methods exploit the convolutional neural network (CNN) architecture to learn a complex non-linear function that maps an input image to an output pose. While such end-to-end methods have demonstrated some success, they have not achieved similar accuracy as geometry-based solutions (e.g., those that optimise pose from a correspondence set). Moreover, recent work~\cite{Sattler2019understanding} suggests that ``\emph{absolute pose regression approaches are more closely related to approximate pose estimation via image retrieval}'', thus they may not generalise well in practice.

\paragraph{Feature learning methods}

Instead of handcrafting descriptors to be robust against varying kinds of distortion so that the distances between them can be used reliably to indicate keypoint matching, some methods resort to machine learning to identify keypoints detected from different views, such as Fern~\cite{Ozuysal2009fast}. It uses a Naive Bayes classifier to recognize keypoints based on a binary descriptor similar to BRIEF~\cite{Calonder2010brief}, which is produced by pixel intensity comparisons. 


While the keypoint matching problem can be solved using machine learning, deep CNN-based feature learning methods typically fix the 2D-3D keypoint associations and learn to predict the image locations of each corresponding 3D keypoint such as~\cite{Peng2019pvnet, Pavlakos20176,Tekin2018real}. They mainly differ in model architecture and the choice of keypoints. For instance, \cite{Pavlakos20176} uses semantic keypoints while \cite{Tekin2018real} chooses the vertices of the 3D bounding box of an object. In our space-borne scenario, objects are typically not occluded and have relatively rich texture. As a result, we opt for object surface keypoints in order to better relate them to strong visual features. 

Another common characteristic of aforementioned CNN-based methods is that, in spite of their various designs of architecture, they all gradually transform the feature maps of the input image from high-resolution representations to low-resolution representations, and recover them to high-resolution representations again at a later stage. Recent research has shown the importance of maintaining a high-resolution representation during the whole process in various tasks including object detection and human pose estimation~\cite{Sun2019deep,Sun2019high}. Specifically, the High-Resolution Net (HRNet) \cite{Sun2019deep} which maintains a high-resolution representation while exchanging information across the parallel multi-resolution subnetworks throughout the whole process, as illustrated in Figure~\ref{fig:hrnet}, produces heatmaps of landmarks with superior spatial precision. To achieve state-of-the-art accuracy in satellite pose estimation, in our framework we use the HRNet for predicting the locations of 2D landmarks in each image.

\subsection{Spacecraft pose estimation}

Monocular spacecraft pose estimation techniques usually adopt a model-based approach. For example,~\cite{D2014pose, Sharma2018robust} first preprocess the images and use feature detectors to identify prominent features such as line segments and basic geometric shapes. Search algorithms are then used to find the right matches between the detected features and the 3D structure. Lastly poses are computed using PnP solvers such as EPnP~\cite{Lepetit2009epnp} and are further refined using optimisation techniques. As summarised in Section~\ref{sec:intro}, our approach also generates 2D-3D correspondences; however, we use a trained deep network to regress the coordinates of 2D landmarks.

The Spacecraft Pose Network (SPN)~\cite{Sharma2019pose} is the seminal work on the SPEED. SPN uses a hybrid of classification and regression neural networks for the pose estimation problem. To perform classification, SPN discretises the 3D rotation group SO(3) into $m$ uniformly distributed base rotations. SPN first predicts the bounding box of the satellite in the image with an object detection sub-network. Then, a classification sub-network retrieves the $n$ most relevant base rotations from the feature map of the detected object. This regression sub-network learns a set of weights and outputs the predicted rotation as a weighted average of the $n$ base rotations. Lastly, SPN solves the relative translation of the satellite utilising constraints from the predicted bounding box and rotation. 

For a more comprehensive survey of spacecraft pose estimation, we refer the reader to~\cite{reviewPAS19}.

\section{Methodology}\label{sec:methodology}

Figure~\ref{fig:pipeline} describes the overall pipeline of our methodology, which consists of several main modules: using a small subset of manually chosen training images (9 images were chosen), we first reconstruct a 3D structure of the satellite with a number of manually chosen landmarks ($11$ was chosen in our implementation) via multi-view triangulation (recall that the training images were supplied with ground truth poses). An object detection network is then used to predict the 2D bounding box of the satellite in the input image. The bounded subimage is then subjected to a landmark regression network to predict the $11$ landmark image positions. Finally, we solve for the poses using the predicted 2D-3D correspondences. Details of the main steps are described in the rest of this section. Our code is available in~\cite{spe}.

\subsection{Multi-view triangulation}

\begin{figure}
    \centering
    \includegraphics[width = \linewidth]{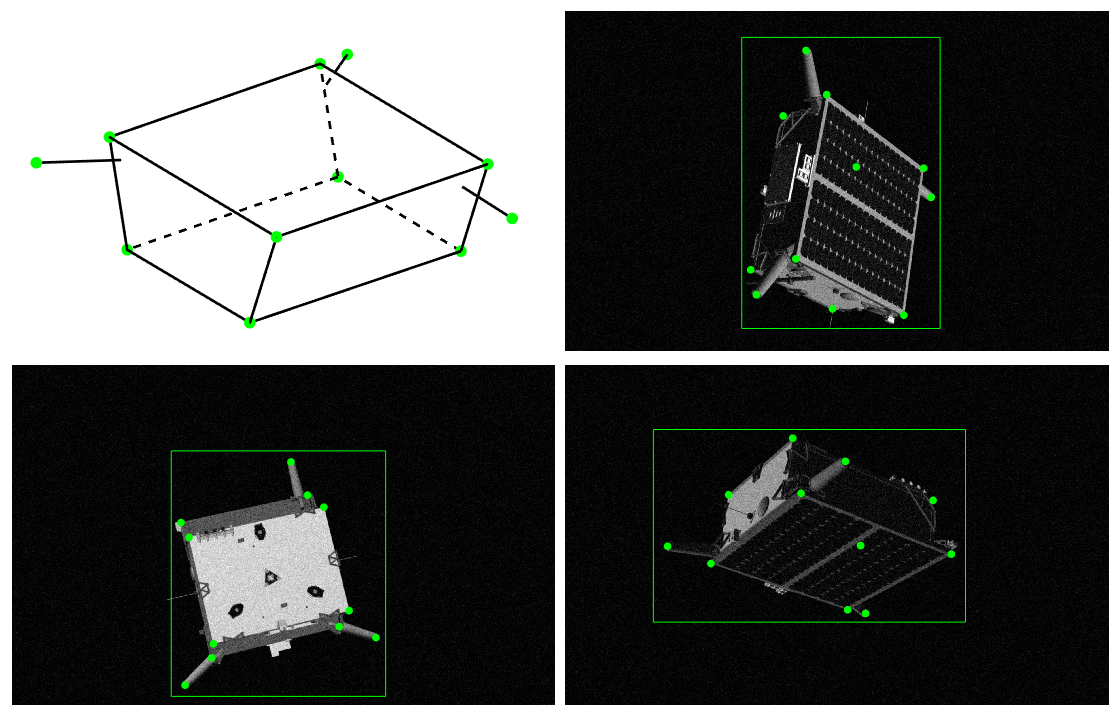}
    \caption{The reconstructed 3D model with 11 landmarks and 3 examples of the bounding boxes determined by the projected 2D landmarks.}
    \label{fig:3dmodel}
\end{figure}

We represent the structure of the object with a small number $N$ of 3D landmarks $\{\bx_i\}_{i=1}^N$ such that they correspond to strong visual features in the images. For the satellite, we select its eight corners plus the centres of the ends of its three antennas, which make a total of $N=11$ landmarks. We use multi-view triangulation to reconstruct the 3D structure. To generate the input for triangulation (i.e., 2D-3D correspondences), we manually match every 3D point with 2D corresponding points over a few handpicked close-up images from the training set. Let $\bz_{i,j}$ denote the 2D coordinates of the $i$-th landmark obtained from the $j$-th image, the 3D landmarks $\{\bx_i\}$ are reconstructed by solving the following objective\footnote{We used the routine \Verb:triangulateMultiview: in MATLAB.}:
\begin{equation}
    \min_{\{\bx_i\}_{i=1}^N} \sum_{i,j} || \bz_{i,j} - \pi_{\bT_j^*}(\bx_i) ||_2^2\,,
\end{equation}
where $\bT_j^*$ is the ground truth pose of image $j$ and $\pi_\mathbf{T}$ is the projective transformation of a structural point into the image plane with pose $\mathbf{T}$ and known camera intrinsics. Figure~\ref{fig:3dmodel} shows the 11 selected 3D landmarks and the reconstructed model as a wireframe.

\subsection{Object detection}\label{sec:obj_detect}



Our pipeline starts by obtaining a bounding box of the object in the image. The aforementioned set of structural landmarks $\{\bx_i\}$ facilitates object detection since the convex hull of their 2D matches $\{\bz_i\}$ covers almost the whole object in any image. Hence a simple but effective method to obtain the ground truth bounding box is to slightly relax the (axis-aligned) minimum rectangle that encloses all $\bz_i$, as shown in Figure~\ref{fig:3dmodel}. We use this method for the training images for which we obtain the ground truth 2D landmarks $\{\bz_i^*\}$ by projecting $\{\bx_i\}$ to the image plane with the ground truth camera pose $\bT^*$, \ie, 
\begin{equation}
    \bz_i^* = \pi_{\bT^*}(\bx_i), \; i = 1,...,N \,.
\end{equation}

For the testing images, we train an object detection model to predict the bounding boxes. We use an off-the-shelf object detection model described in~\cite{Sun2019high}, which applies an HRNet as backbone in the Faster-RCNN \cite{Ren2015faster} framework. The HRNet backbone is initialised with a pretrained model HRNet-W18-C\footnote{The pretrained model was downloaded from~\cite{pretrained}.} \cite{Sun2019high}. We train the detection model on the MMDetection platform~\cite{Chen2019mmdetection} and follow the training settings as in~\cite{Sun2019high}.

\subsection{Landmark regression}\label{sec:regression}

Each training image is coupled with a bounding box and a set of ground truth 2D landmarks $\{\bz_i^*\}$ as described in Section~\ref{sec:obj_detect}. We use these labels to supervise the training of a regression model to predict the 2D landmarks in the testing images.
Additionally, to handle images that only capture partial object, we label the visibility $v_i$ of each 2D landmark $\bz_i^*$ of each image in the training set where
\begin{equation}
    v_i = \begin{cases}
    1 \quad \text{if $\bz_i^*$ is inside image frame,} \\
    0 \quad \text{otherwise.}
    \end{cases}
\end{equation}

We used the HRNet as described in~\cite{Sun2019deep} to regress the 2D landmark locations. Specifically we used pose-hrnet-w32~\cite{hrnet_arch} for our architecture (Figure~\ref{fig:hrnet}), which has 32 channels in the highest resolution feature maps. The output of the model is a tensor of 11 heatmaps; one for each 3D landmark. Because of this model-designed one-to-one association between 3D landmarks and heatmaps, the model solely has to learn the image location of each 3D landmark but not the heatmap-3D landmark associations.



To increase the prediction accuracy as well as robustness against the Earth background, we crop each image with their bounding boxes and resize them to fit the input window of the regression model. We conduct this process in both the training and the testing phase. For the later, we predict the bounding boxes of the testing images with the object detection model. Because HRNet maintains a high-resolution representation, it is able to produce high-resolution heatmaps with superior spatial accuracy. To leverage this characteristic of HRNet, we increased the size of the input window as well as the size of the output heatmaps to $768\times 768$ from the default $256 \times 256$.

We train the model from scratch by minimising the following loss:
\begin{equation}
    \ell = \frac{1}{N}\sum_{i=1}^N v_i(h(\bz_i) - h(\bz_i^*))^2\,,
\end{equation}
\ie, the mean squared errors between the predicted heatmaps $h(\bz_i)$ and ground truth heatmaps $h(\bz_i^*)$ of the visible landmarks in each image. The notation $h(\cdot)$ denotes a heatmap representation of a 2D point. We generate the ground truth heatmaps as 2D normal distributions with means equal to the ground truth locations of each landmark, and standard deviations of 1-pixel. The loss function $\ell$ is defined based on a single image. In a mini batch, $\ell$ is simply averaged. The model is trained for 180 epochs with the Adam optimizer \cite{Kingma2014adam}. Other training setup is adopted from \cite{Sun2019deep}.

\subsection{Pose estimation}\label{sec:pose}

The final step in our pipeline is to estimate the pose $\mathbf{T} \in \text{SE}(3)$ for a test image given the predicted 2D-3D correspondences $\{(\mathbf{z}_i, \mathbf{x}_i)\}$ as described in Section~\ref{sec:regression}. We estimate $\mathbf{T}$ by solving the robust non-linear least-squares problem
\begin{equation}\label{eq:ls}
    \min_{\mathbf{T}} \sum_i L_\delta( r_i(\bT) )
\end{equation}
with residuals 
\begin{equation}\label{eq:res}
    r_i(\bT) = \left\| \mathbf{z}_i - \pi_\mathbf{T}(\mathbf{x}_i) \right\|_2,
\end{equation}
\noindent and subject to cheirality constraints. $L_\delta: \mathbb{R} \rightarrow [0, \infty)$ is the Huber loss 
\begin{equation}
L_\delta(r) = \begin{cases}
\dfrac{r^2}{2} & \; \text{if } |r|\leq \delta\\
\delta|r|- \dfrac{\delta^2}{2}& \; \text{otherwise. }
\end{cases}
\end{equation}


We use Levenberg-Marquardt (LM) to solve Eq.~\eqref{eq:ls}; we called LMPE to our C++ implementation with the Ceres Solver~\cite{ceres-solver}. We can run LMPE after setting $\delta$ and choosing an initial linearisation point $\bT_0$; however, picking a value for $\delta$, and potential outlying correspondences could impact on producing an accurate estimation. Instead, we propose a Simulated Annealing scheme (SA-LMPE) as depicted in Algorithm~\ref{alg:sa-lmpe} to progressively adjust $\delta$ and remove potential outlying correspondences. A correspondence $(\bz_i,\bx_i)$ is regarded as an outlier if 
\begin{equation}
    r_i(\bT^*)  > \epsilon
\end{equation}
\noindent for a threshold $\epsilon$, and the ground truth pose $\bT^*$. In practice, we use the residual with respect to the current pose $r_i(\bT_{t+1})$ to indicate potential outliers for removal. 


\begin{algorithm}
\caption{SA-LMPE.}
\label{alg:sa-lmpe}
\begin{algorithmic}[1]
\REQUIRE 2D-3D matches $H_0 := \{(\bz_i, \bx_i)\}$,  initial pose $\bT_0$, initial values for $\delta$ and $\epsilon$, cooling parameters $\delta_{\text{min}}, \epsilon_{\text{min}}>0,\, 0<\lambda_\delta, \lambda_\epsilon\leq 1$, and number of iterations $t_\text{max}$.
\STATE $t \leftarrow 0$.
\WHILE{$t<t_\text{max}$}
\STATE $\bT_{t+1} \leftarrow \; \text{LMPE} (H_t, \bT_{t}, \delta)$.\label{step:lmpe}
\STATE $H_{t+1} \leftarrow \;  \left\{ (\bz_i, \bx_i) \in H_t\,|\, r_i(\bT_{t+1})  \leq \epsilon \right\} $.\label{step:outrem}
\STATE $\delta \leftarrow \max(\delta_{\text{min}},\,\lambda_\delta \delta)$.\label{step:cooling_delta}
\STATE $\epsilon \leftarrow \max(\epsilon_{\text{min}}, \lambda_\epsilon \epsilon)$.\label{step:cooling_eps}
\STATE $t \leftarrow t+1$.
\ENDWHILE
\RETURN $T_{t}$.
\end{algorithmic}
\end{algorithm}

\begin{table*}[h]
    \begin{center}
    \begin{tabular}{c|c|c}
       \hline
       Metric  &  \makecell{SPN \cite{Sharma2019pose} \\ (on test set)} & \makecell{Ours \\ (on training set CV)} \\
       \hline\hline
       Mean IOU  &  0.8582 & 0.9534 \\
       Median IOU & 0.8908 & 0.9634 \\
       Mean $E_R$ (degree)  &  8.4254 & 0.7277\\
       Median $E_R$ (degree) & 7.0689 & 0.5214 \\
       Mean $E_T$ (metre)  &  N/A & 0.0359\\
       Median $E_T$ (metre) & N/A & 0.0147 \\
       Mean $|\bm{t}^* - \bm{t}|$ (metre) &  [0.0550,\quad 0.0460,\quad 0.7800] & [0.0040,\quad 0.0040,\quad 0.0346]\\
       Median $|\bm{t}^* - \bm{t}|$ (metre) & [0.0240,\quad 0.0210,\quad 0.4960] & [0.0031,\quad 0.0030,\quad 0.0134]\\
       \hline
    \end{tabular}
    \end{center}
    \caption{Performance comparison between the SPN and the proposed method.}
    \label{tab:result}
\end{table*}

There is a virtuous circle in our annealing process: an accurate pose will help on carefully removing potential  outliers (Line~\ref{step:outrem}), while lesser outlying corrupted data will produce a more accurate estimation (Line~\ref{step:lmpe}). Thus, initial $\delta$ and $\epsilon$ values progressively ``cool down'' (Step~\ref{step:cooling_delta} and Step~\ref{step:cooling_eps}), until reaching minimum predefined values ($\delta_\text{min}, \epsilon_\text{min}$) or a maximum number of iterations $t_\text{max}$.

For the SPEED images, we obtained the initial pose $\bT_0$ in Algorithm~\ref{alg:sa-lmpe} by using a RANSAC fashion PnP solver\footnote{We used the routine \Verb:estimateWorldCameraPose: in MATLAB.} (with kernel P3P~\cite{gao03} and minimal four-points sets) on the predicted correspondences.



\section{Evaluation}

In this section we report the evaluation metrics and experimental results of our methodology.

\subsection{Metrics}

We evaluate the estimated pose of each image using a rotation error $E_R$ and a translation error $E_T$. Let $q^*$ and $q$ denote the rotation quaternion ground truth of an image and its estimation. Analogously, let $\bm{t}^*$ and $\bm{t}$ denote the ground truth and estimated translation vectors of an image. We then define $E_R$ and $E_T$ as
\begin{equation}
    E_R = 2\cos^{-1}\left(|z|\right)\,,
\end{equation}
where $z$ is the real part of the Hamilton product between $q^*$ and the conjugate of $q$, \ie, 
$
    z+\bm{c} = q^*\,\text{conj}(q)\,,
$
where $\bm{c}$ is the vector part of the Hamilton product and
\begin{equation}
    E_T = ||\bm{t}^* - \bm{t}||_2\,.
\end{equation}


We report our object detection results via the Intersection Over Union (IOU) score based on CV. For each image, its IOU score is the intersection area divided by the union area of the predicted and the ground truth bounding boxes.


We compare against KPEC's participants trough the scores defined in the KPEC: namely the rotation score $S_R$, the translation score $S_T$, and the overall score $S$. $S_R$ is the same as $E_R$ but in radians, 
\begin{equation}
    S_T = \frac{||\bm{t}^* - \bm{t}||_2}{||\bm{t}^*||_2}\,,
\end{equation}
and 
\begin{equation}
    S = S_R + S_T\,.
\end{equation}

\subsection{Experiments}


Since the KPEC withheld the ground truth poses of the test set, we cannot conduct analysis based on the test set other than providing the overall score. Instead, we analysed our method using 6-fold CV over the training set. Specifically, we split the 12,000 training images into 6 groups, and then for each group, we train an object detection (Section~\ref{sec:obj_detect}) model and a landmark regression (Section~\ref{sec:regression}) model with the images in the remaining 5 groups. We test each model with their respective designated group, i.e., the complement of the 5 groups we train the model with. Thus each model is equipped with a disjoint test group so that in total, they cover all 12,000 images in the training dataset. 


Following the above CV procedure, we estimate the pose of every training image. In effect, we predict the image coordinates of every 3D landmark to obtain 2D-3D correspondences from which we obtained an initial pose using RANSAC with a PnP kernel, which we finally refine with Algorithm~\ref{alg:sa-lmpe}. We make clear that we invoke Algorithm~\ref{alg:sa-lmpe} with all predicted 2D-3D correspondences and not with the consensus set after RANSAC. 

For the test set, we exploit the advantage of ensemble methods since we have 6 trained models resulted from the 6-fold CV. We average the 6 heatmaps predicted by the 6 trained models for each landmark and each test image before we obtain the final 2D landmark coordinates. The rest of the precedure is the same as described in Section~\ref{sec:pose}.



\begin{figure*}[h]
\centering
    \subfloat[$S_R$ ]{\includegraphics[width=.32\linewidth]{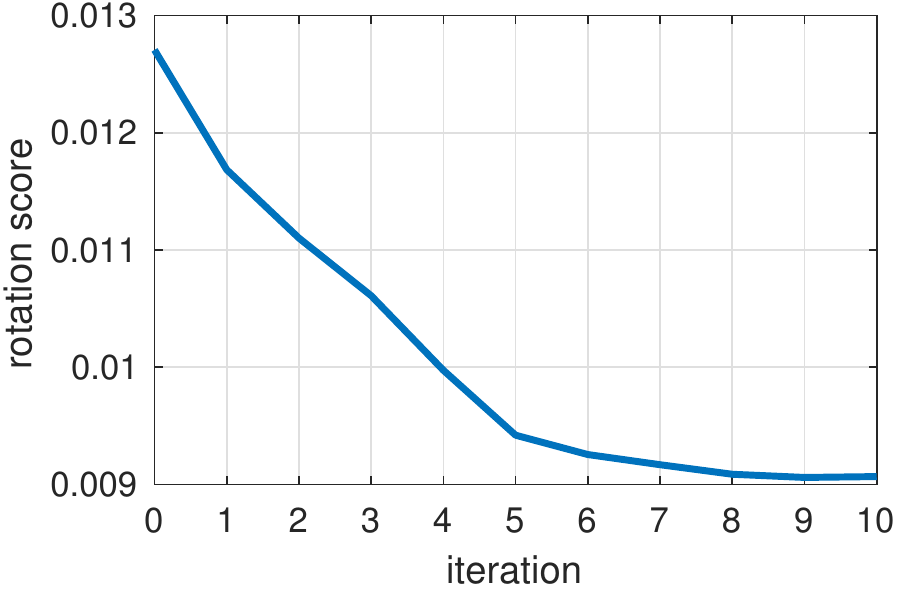}}\;
    \subfloat[$S_T$]{\includegraphics[width=.32\linewidth]{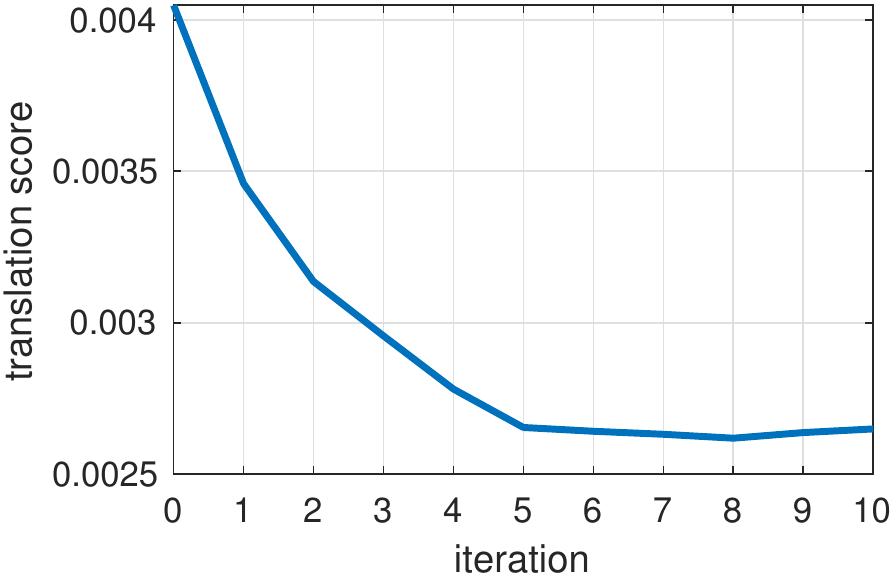}}\;
    \subfloat[$S$]{\includegraphics[width=.32\linewidth]{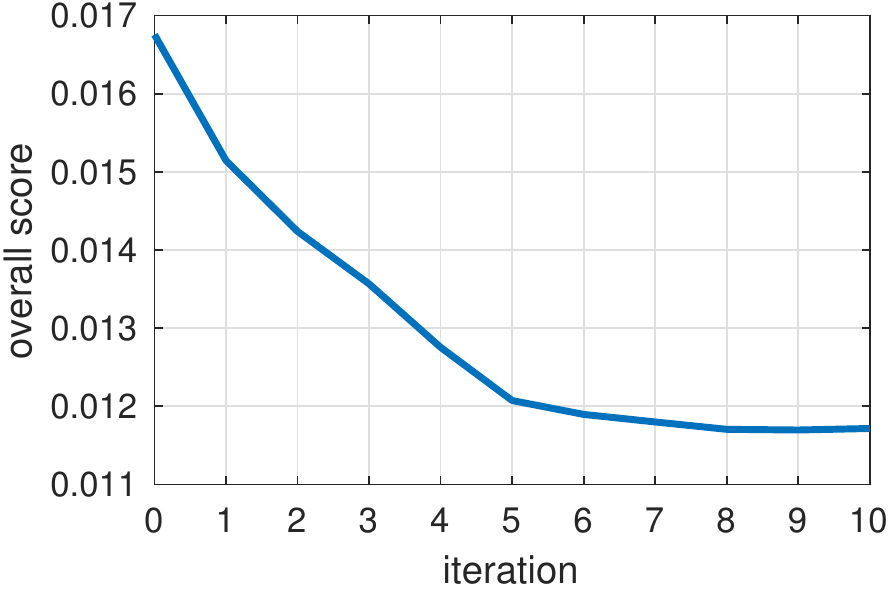}}
    \caption{Score evolution for SA-LMPE over all images. }
    \label{fig:error_evolution}
\end{figure*}

\begin{figure}
    \centering
    \subfloat[RANSAC ]{\includegraphics[width=.45\linewidth]{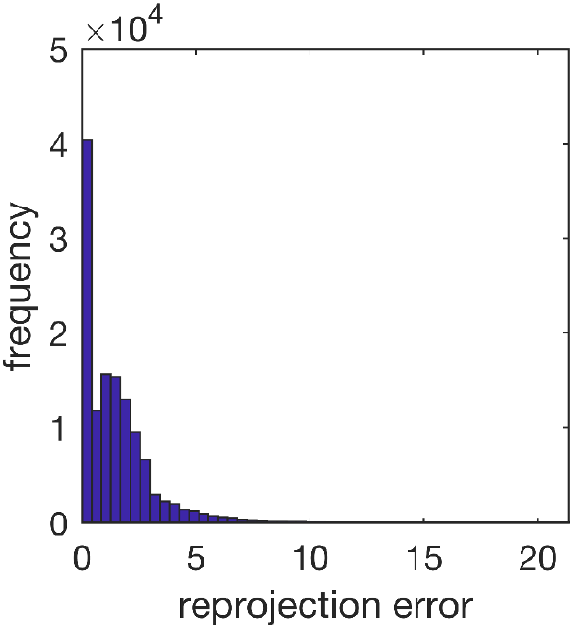}} \;\;\;
    \subfloat[SA-LMPE]{\includegraphics[width=.45\linewidth]{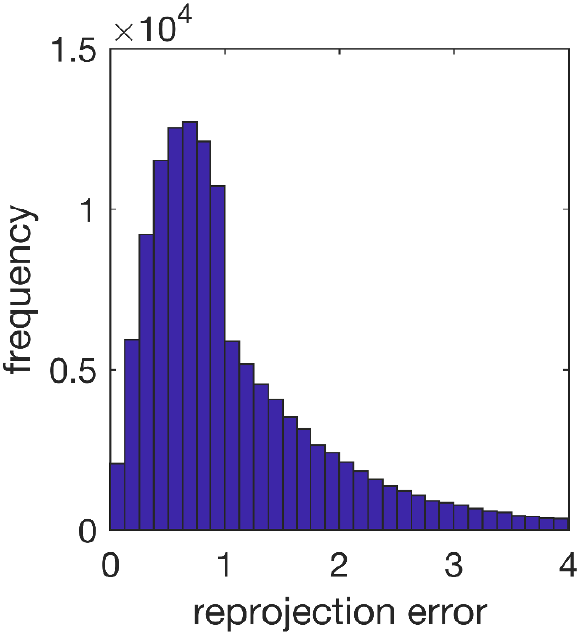}}
    
    \caption{Histograms of reprojection errors of all landmarks from all images with respect to (a) the initial poses obtained by RANSAC, and (b) the final poses after the SA-LMPE refinement. For better visualisation of the RANSAC histogram, we truncated its long tail by removing the $1\%$ largest errors. }
    \label{fig:res_hist_init}
\end{figure}

\subsection{Results}


We first compare against SPN \cite{Sharma2019pose}; Table~\ref{tab:result} report the performance results. Our proposed method achieves superior performances in both object detection and pose estimation. Both our rotational and translational errors are at least one degree of magnitude smaller than SPN.

In terms of the KPEC scores, our average overall score of the training set based on a 6-fold CV is 0.0117. To investigate the effect of the pose refinement, Figure~\ref{fig:error_evolution} displays the evolution of average scores during the refinement process while Figure~\ref{fig:res_hist_init} provides a comparison of reprojection residuals before and after the refinement. Based on the error distribution of the initial poses in Figure~\ref{fig:res_hist_init}(a), we set $\delta=5$ and $\epsilon=50$ to initialise SA-LMPE. For the cooling parameters we take $\delta_\text{min}=1$, $\epsilon_\text{min}=4$, and $\lambda_\delta=\lambda_\epsilon=0.7$. We set the maximal number of iterations $t_\text{max}=10$. SA-LMPE removed $8495$ potential outliers in total which is equivalent to approximately $0.7$ outliers per image. As shown in Figure~\ref{fig:error_evolution}, the pose refinement improves the average overall score $S$ from $0.0167$ to $0.0117$. 

\begin{table}[]
    \begin{center}
    \begin{tabular}{c|c|c}
        \hline
        Rank & Participant Name & Score \\
        \hline\hline
        1 & UniAdelaide & 0.0094 \\
        2 & EPFL\_cvlab & 0.0215 \\
        3 & pedro\_fairspace & 0.0571 \\
        4 & stanford\_slab & 0.0626 \\ 
        5 & Team\_Platypus & 0.0703 \\
        6 & motokimura1 & 0.0758 \\ 
        7 & Magpies & 0.1393 \\
        8 & GabrielA & 0.2423\\
        9 & stainsby & 0.3711\\
        10 & VSI\_Feeney & 0.4658 \\
        \hline
    \end{tabular} 
    \end{center}
    \caption{Top 10 scores of KPEC.}
    \label{tab:top10}
\end{table}

\begin{figure*}
    \centering
    \includegraphics[width = 0.94\linewidth]{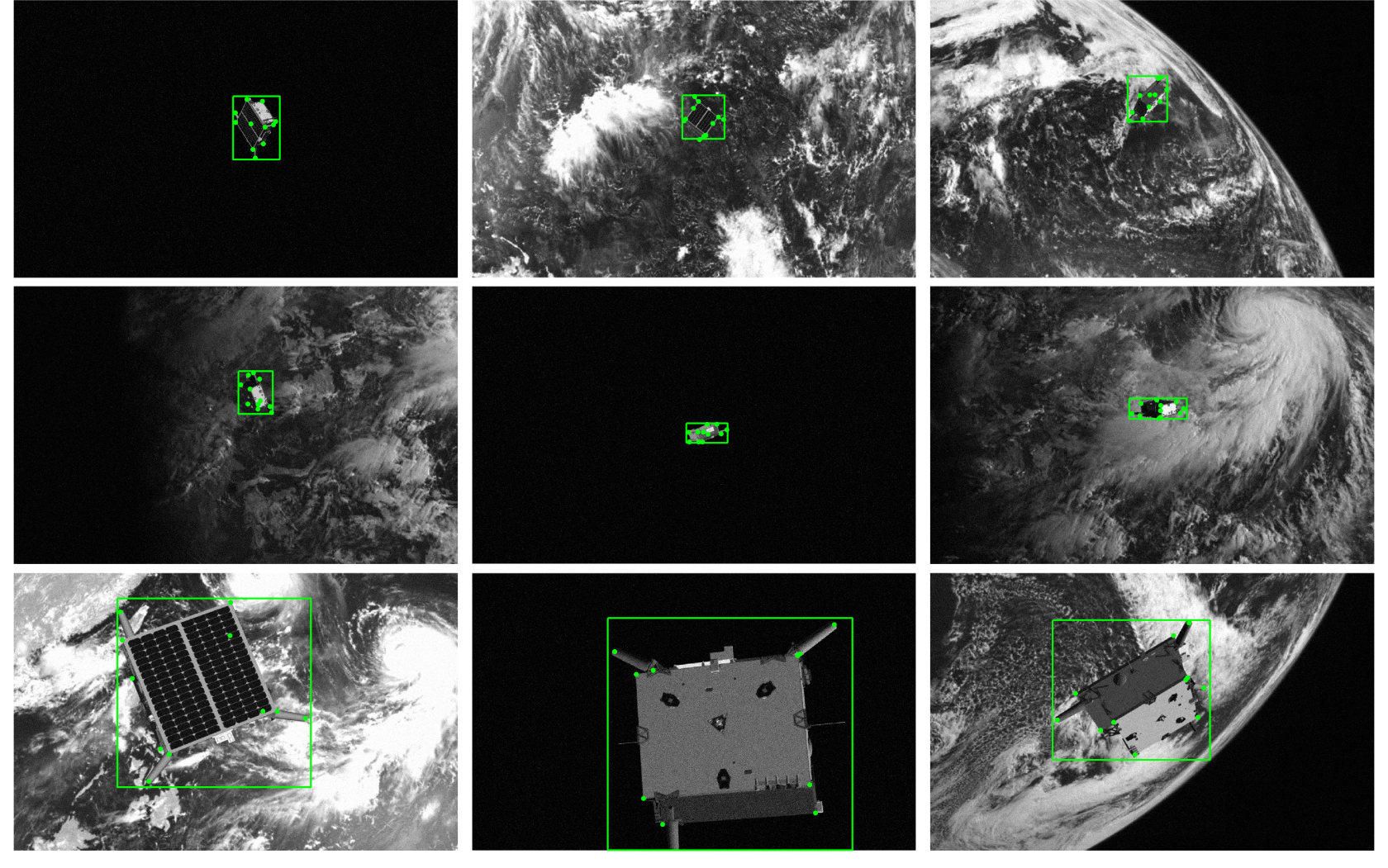}
    \caption{A montage of random test images with the predicted bounding boxes of the satellite and the estimated 2D landmarks.}
    \label{fig:testbbox}
    \vspace{0.6cm}
    \centering
    \includegraphics[width = 0.94\linewidth]{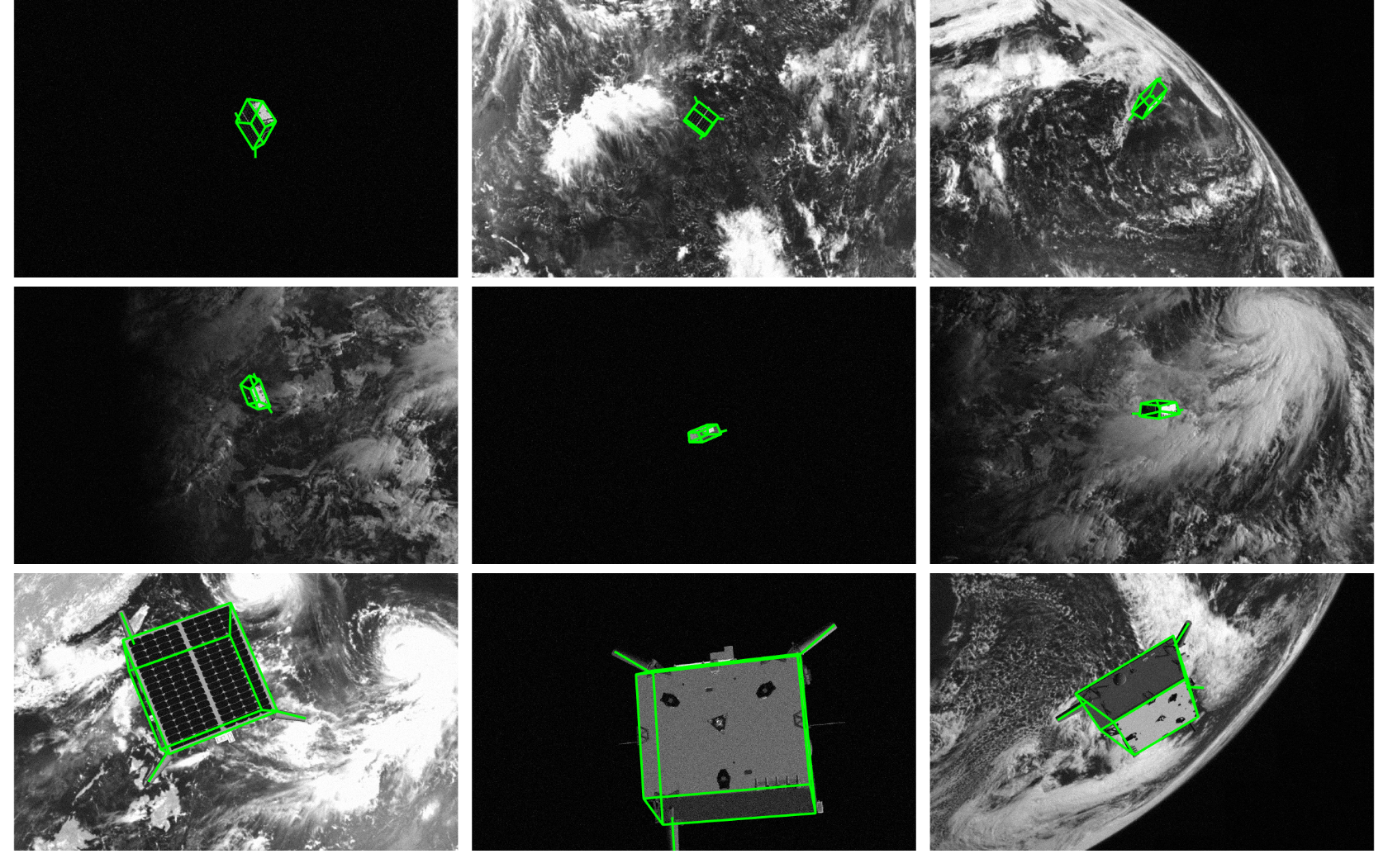}
    \caption{A montage of the same test images in Figure~\ref{fig:testbbox} with the predicted poses shown as green wireframes.}
    \label{fig:test_result}
\end{figure*}

Our overall score of the test set is 0.0094 which is slightly better than the training set CV 0.0117, thanks to the benefits from the ensemble of 6 models. Table~\ref{tab:top10} provides the top 10 scores in KPEC. We provide Figure~\ref{fig:testbbox} and~\ref{fig:test_result} for visual inspection of object detection, landmark regression and pose estimation results on a sample of the test set. Note that we did not cherry-pick the images from testing results -they were selected at random. Visual inspection indicates high accuracy of our approach even with images that have very small object size.

\section{Conclusion}
We propose a monocular pose estimation framework for space-borne objects such as satellite. Our framework exploits the strength of deep neural networks in feature learning and geometric optimisation in robust fitting. In particular, the high-resolution representation of images used in HRNet enables accurate predictions of 2D landmarks while the SA-LMPE algorithm allows further removal of inaccurate predictions and refinement of poses. 


Our approach won the first place in the the KPEC. Our CV-based evaluation also indicates our method significantly outperforms previous work on the SPEED benchmark.

\subsection*{Acknowledgement}

This work was jointly supported by ARC project LP160100495 and the Australian Institute for Machine Learning.

\clearpage

{\small
\bibliographystyle{ieee}
\bibliography{egbib}
}

\end{document}